# Biomimetic Machine Learning approach for prediction of mechanical properties of Additive Friction Stir Deposited Aluminum alloys based walled structures

Akshansh Mishra[1,*]

[1]School of Industrial and Information Engineering, Politecnico Di Milano, Milan, Italy

**Abstract:** This study presents a novel approach to predicting mechanical properties of Additive Friction Stir Deposited (AFSD) aluminum alloy walled structures using biomimetic machine learning. The research combines numerical modeling of the AFSD process with genetic algorithm-optimized machine learning models to predict von Mises stress and logarithmic strain. Finite element analysis was employed to simulate the AFSD process for five aluminum alloys: AA2024, AA5083, AA5086, AA7075, and AA6061, capturing complex thermal and mechanical interactions. A dataset of 200 samples was generated from these simulations. Subsequently, Decision Tree (DT) and Random Forest (RF) regression models, optimized using genetic algorithms, were developed to predict key mechanical properties. The GA-RF model demonstrated superior performance in predicting both von Mises stress ($R^2 = 0.9676$) and logarithmic strain ($R^2 = 0.7201$). This innovative approach provides a powerful tool for understanding and optimizing the AFSD process across multiple aluminum alloys, offering insights into material behavior under various process parameters.

## 1. Introduction

Additive Friction Stir Deposition (AFSD) is a friction stir-based additive manufacturing process that involves the layer-by-layer deposition of material using feedstock, substrate, and tools [1-5]. This procedure is based on friction stir welding (FSW), in which the material experiences intense thermoplastic deformation without melting, resulting in a thin, equiaxed microstructure in the finished product. There are three main variants of friction stir-based deposition additive manufacturing technologies: Friction Surface Deposition Additive Manufacturing (FSD-AM), Friction Extrusion Additive Manufacturing (FEAM), and Additive Friction Stir Deposition (AFSD). Friction Surface Deposition Additive Manufacturing (FSD-AM) involves using metal rods as feedstock. These rods are fixed to a spindle that rotates and presses down, generating heat through friction stirring. The plasticized material is then layered onto the substrate to form the additive component. As the spindle moves along the process trajectory, the component is formed, although the material experiences unconstrained expansion in both radial and axial directions, often resulting in curled edges around the rod. Friction Extrusion Additive Manufacturing (FEAM) uses metal rods which are transformed into a plastic state through friction with a rotating die driven by axial force. The plasticized metal is then extruded from the die outlet and fills the gap between the substrate and the tool, forming the component as the spindle moves. This method, however, tends to produce a poorly bonded layer due to the frictional interaction between the feedstock and rotating die. Additive Friction Stir Deposition (AFSD), which is shoulder-assisted, generally employs rods, wires, or powders as raw materials [6-10]. These materials are introduced into a hollow, non-consumable tool and, under extrusion, friction, and stirring effects, become thermoplasticized and migrate downwards to the substrate. The mechanical mixing of the softened substrate and plasticized raw material creates a robust bond, followed by component formation as the spindle traverses a predefined path. Compared to the other two techniques, AFSD offers more precise control over material flow and forming morphology. Several essential parameters influence the AFSD process, such as tool rotating speed, feed rate, and layer height. The rate of heat generation is primarily determined by the tool's spinning speed, whereas the feed rate or axial force governs the rate of material deposition. The tool traverse velocity influences the geographical distribution of heat, whereas layer height defines the distance between the tool and the substrate. During the operation, material flow is driven by extrusion and shearing in the transition zone beneath the feedstock rod, with the tool's stirring action also playing an important role [11-14]. The thermal evolution of AFSD is defined by heat created by friction and plastic deformation, with the internal temperature distribution influencing material flow.

Machine learning integration into the AFSD process is motivated by a number of key needs and opportunities to improve efficiency, precision, and overall performance. Machine learning can help handle the challenges of AFSD, such as controlling a large number of process parameters, anticipating material behavior, and assuring product quality. AFSD involves a number of interdependent factors, including tool rotation speed, feed rate, and axial force, all of which influence material flow, temperature distribution, and mechanical properties of the deposited layers. Traditional trial-and-error approaches to optimizing these parameters are time-consuming and expensive. Machine learning algorithms can examine massive datasets generated by AFSD processes to more efficiently identify optimal parameter settings, minimizing the need for lengthy experimental runs. Qiao et al. [15] investigated the use of machine learning approaches to optimize process parameters in AFSD for greater component design flexibility and performance. They used support vector machine (SVM), random forest (RF), and artificial neural network (ANN) models to predict the mechanical properties of AFSD-based AA6061 deposition, particularly microhardness and ultimate tensile strength (UTS). Temperature, force, torque, rotation speed, traverse speed, feed rate, and layer thickness were all monitored with a self-developed process-aware kit. The study found that the ANN model was the most accurate, with a $R^2$ of 0.9998, and low errors (MAE of 0.0050 and RMSE of 0.0063). The study also revealed feed rate and layer thickness as major factors impacting mechanical properties, with contributions of 24.8%/24.1% and 25.6%/26.6%. Zhu et al. [16] suggested a novel explainable artificial intelligence strategy that combines Bayesian learning and physics-based surrogate models. They created a physics-informed, data-driven model that can accurately forecast temperature distribution during AFSD by calibrating and updating these models using machine learning on in-situ monitoring data. The usefulness of this approach was demonstrated by the AFSD of an Al-Mg-Si alloy, which yielded fast and accurate temperature forecasts with minimum physics simulation runs and in-situ measurements. Shi et al. [17] introduced a physics-informed machine learning approach named AFSD-Nets which combines heat generation and heat transfer effects to predict temperature profiles. AFSD-Nets incorporates customized neural network approximators to model the coupled temperature evolution of the tool and the build during multi-layer material deposition. The study demonstrated that AFSD-Nets could accurately predict temperature evolution during the AFSD process, as validated by a comparison between predictions and actual measurements.

The present study intends to build a unique biomimetic machine learning approach for predicting the mechanical properties of Additive Friction Stir Deposited (AFSD) aluminum alloy walled structures in light of these opportunities and challenges. Specifically, AA2024, AA5083, AA5086, AA7075, and AA6061 are the five aluminum alloys on which the study concentrates. To anticipate von Mises stress and logarithmic strain, the method integrates genetic algorithm-optimized machine learning models with finite element analysis of the AFSD process.

## 2. Problem Statement

AFSD is a potential manufacturing method for creating aluminum alloy structures. However, the complex relationship of thermal and mechanical processes during AFSD makes it difficult to anticipate and regulate the mechanical properties of the finished product. Traditional experimental methods for optimizing process parameters and predicting material behavior are time consuming and expensive. In addition, the variety in material qualities among aluminum alloys (AA2024, AA5083, AA5086, AA7075, and AA6061) worsens this problem. There is a need for an efficient and accurate approach to predicting critical mechanical parameters such as von Mises stress and logarithmic strain for AFSD-manufactured components made from diverse aluminum alloys. This research aims to address this gap by developing a novel biomimetic machine learning approach that combines numerical modeling with genetic algorithm-optimized predictive models, potentially revolutionizing process optimization and quality control in AFSD manufacturing.

## 3. Material and Methods

Figure 1 shows the basic setup for our simulation work. First, we have to create the parts representing the substrate and the material to deposited in the substrate. Then we create the material properties like density, specific heat, thermal conductivity, elastic and plastic properties which are temperature dependent, and these properties are further assigned to our substrate and our material to be deposited.

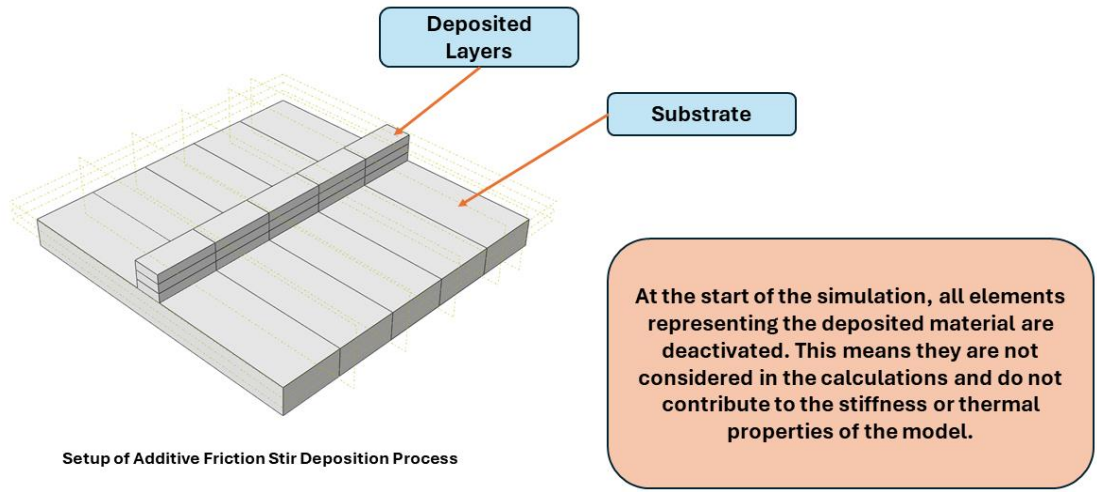

Deposited Layers

Substrate

At the start of the simulation, all elements representing the deposited material are deactivated. This means they are not considered in the calculations and do not contribute to the stiffness or thermal properties of the model.

Setup of Additive Friction Stir Deposition Process

**Figure 1.** Schematic representation of AFSD process

A step represents a specific phase or period within the analysis, during which specific loads, boundary conditions, and analysis procedures are applied. For a coupled temperature-displacement analysis which we will use in our AFSD numerical modelling, a coupled temp-displacement step is utilized. This type of step allows for the simultaneous analysis of both thermal and structural behavior, capturing the interaction between temperature changes and resulting deformations or stresses. Figure 2 represents the elements of the top deposited layer for which the steps will be created. Within these steps, we can specify details like the duration of the step, the loading conditions, the boundary conditions, and the analysis procedure to be used. Abaqus offers various analysis procedures, such as static, dynamic, or explicit, each catering to different simulation requirements.

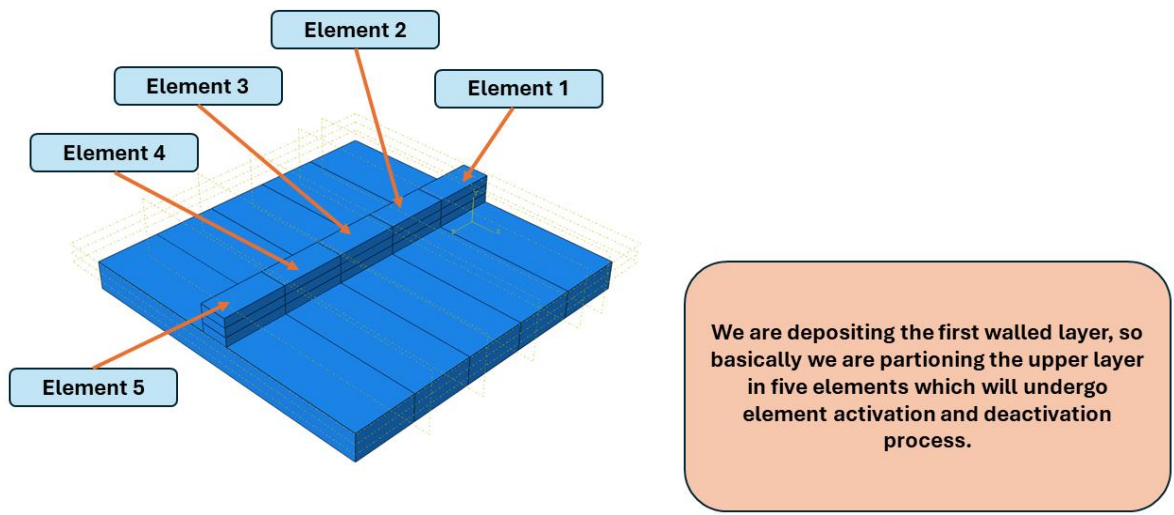

**Figure 2.** Step creation process for the top layer to be deposited

Model change, convection, and radiation are three different interaction types that must be created in order to accurately simulate the complex physical processes involved in Additive Friction Stir Deposition (AFSD) processing shown in Figure 3. In order to accurately represent the additive nature of the AFSD process, model change interactions are necessary to simulate the sequential deposition of material layers. The model can accurately depict the build's changing shape and thermal history by layering on activations. To simulate the heat transfer between the deposited material, the tool, and the surrounding environment, convection interactions are required. This is essential for predicting cooling rates, temperature gradients, and possible flaws like distortions and residual stresses.

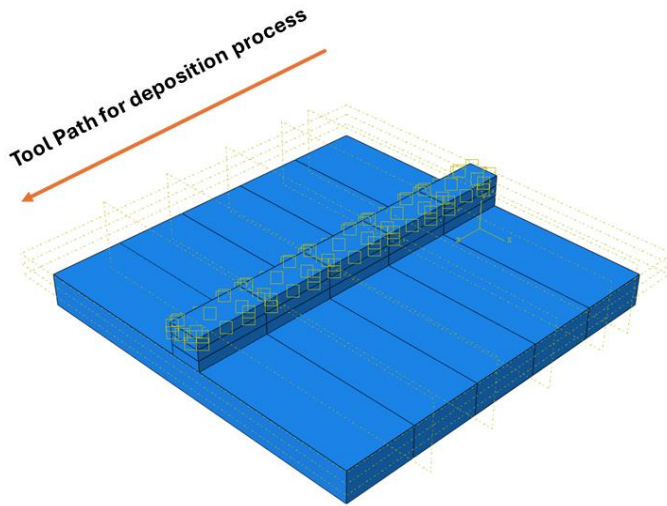

**Three types of interactions are need to be created:**
- **Model Change**
- **Convection**
- **Radiation**

As the simulation progresses, elements representing the deposited material are sequentially activated based on the toolpath and deposition parameters. This simulates the gradual build-up of the part layer by layer.

**Figure 3.** Creation of the interaction for the specified tool path deposition

The loading parameters we need to apply in Abaqus for the Additive Friction Stir Deposition (AFSD) process - heat source, pressure, shear longitude, and shear rotational - are all relevant and important for accurately simulating the process shown in Figure 4.

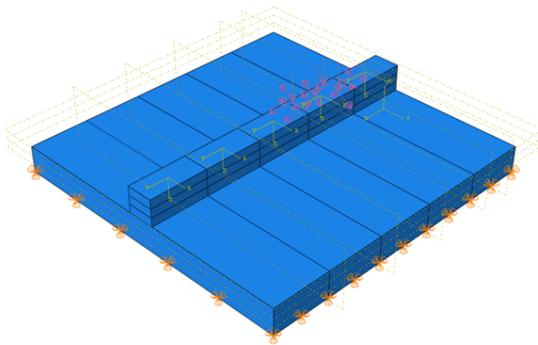

**Four types of loading conditions are need to be created:**
- **Heat Source**
- **Pressure**
- **Shear in longitudinal direction**
- **Shear due to rotation**

**Figure 4.** Schematic representation of the loading parameters used in AFSD process for depositing walled structures

Heat source is the primary driving force in AFSD. The friction between the rotating tool and the substrate generates significant heat, softening the material and enabling deposition. The downward force exerted by the tool is essential for consolidating the deposited material and ensuring a strong bond with the substrate. Pressure also influences the material flow and the shape of the deposited bead. Shear longitude refers to the shear force acting in the direction of tool travel. It plays a role in material transport, mixing, and the formation of the characteristic AFSD microstructure. Shear rotational is the shear force acting in the circumferential direction around the tool. It contributes to material flow, stirring, and the development of the final part geometry. Figure 5 shows the outcomes which we can expect at the end of our simulation. AC YIELD represents the accumulated equivalent plastic strain at the end of each increment. This parameter is crucial for understanding the extent of plastic deformation the material undergoes during the AFSD process. It helps in assessing potential areas of material failure due to excessive strain and in optimizing process parameters to achieve desired material properties. GRADT denotes the spatial gradient of temperature. This parameter is important for understanding the temperature distribution and heat flow within the material during AFSD. It can reveal areas of rapid temperature change, which might lead to thermal stresses and affect the microstructure and mechanical properties of the final part. NT11 represents the normal component of stress in the 11-direction (typically along the build direction in AFSD). This parameter helps in evaluating the stress state in the deposited material, especially the residual stresses that can arise due to the thermal and mechanical loading during the AFSD process. PEEQ denotes the equivalent plastic strain, which is a scalar measure of the plastic deformation in the material. Similar to AC YIELD, PEEQ provides insights into the extent of plastic deformation and can help identify regions prone to failure or areas where material properties might be significantly altered. LE represents the logarithmic strain. This parameter can be used to analyze the deformation behavior of the material under large strains, which are often encountered in AFSD due to the high temperatures and severe plastic deformation. HFL stands for Heat Flux Vector, representing the direction and magnitude of heat flow. This parameter is useful for visualizing and understanding the heat transfer mechanisms during AFSD, such as conduction, convection, and radiation. Analyzing the heat flux can help in optimizing process parameters and cooling strategies. The present work focussed on the deposition of similar alloy layers on the similar alloy based substrate.

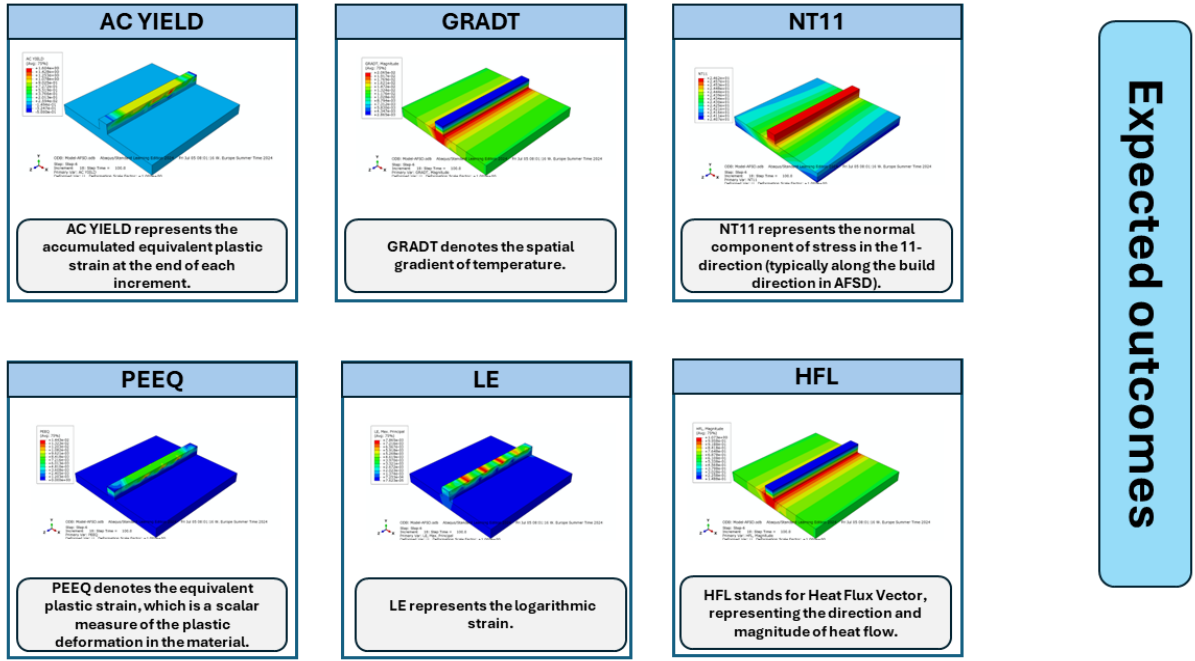

**Figure 5.** Expected results from the numerical modelling

The simulations were conducted using Abaqus finite element software to model the AFSD process for the five aluminum alloys: AA2024, AA5083, AA5086, AA7075, and AA6061 whose properties are shown in Table 1. The input parameters for these simulations included the elastic modulus of the alloys (measured in GPa), specific heat (J/kg·K), shear translation (N), shear rotational (N·m), and heat source (W/m³). These parameters were carefully selected to represent the key physical properties and process conditions that influence the AFSD process. The output parameters of interest were the von Mises stress (MPa) and logarithmic strain (dimensionless), which are critical indicators of the mechanical properties and performance of the deposited structures. From these simulations, a comprehensive dataset of 200 samples was generated, capturing a wide range of process conditions and material behaviors. This dataset was then split into training and testing sets using an 80-20 ratio, with 160 samples used for training the machine learning models and 40 samples reserved for testing and validation. The couple GA-ML algorithms were evaluated on the basis of metric features such as RMSE, MAE and $R^2$ values.

**Table 1.** Properties of the used alloys in present work

| Property | AA2024-T4/T351 | AA5083-H116/H32 | AA5086-H116/H32 | AA7075-T6/T651 | AA6061-T6/T651 |
|---|---|---|---|---|---|
| Elastic Modulus (GPa) | 73.1 | 72 | 70 | 71.7 | 68.9 |
| Density (g/cm³) | 2.78 | 2.66 | 2.66 | 2.81 | 2.70 |
| Specific Heat (J/kg·K) | 875 | 880 | 880 | 960 | 896 |

## 4. Results and Discussion

### 4.1 Numerical Modelling of Additive Friction Stir Deposition process

The element activation and deactivation technique is used in the numerical modeling of the Additive Friction Stir Deposition (AFSD) process to mimic material addition and removal sequentially, correctly capturing the deposited material's changing geometry and temperature history. This technique involves activating parts in stages to simulate the real deposition process, guaranteeing that the growing structure and its thermal and mechanical properties are precisely represented. Model change interactions are an important component of this technique, allowing for the activation of new components matching to newly deposited material and the deactivation of elements to represent removal or non-influence at specified moments. This method is used with a coupled temperature-displacement analysis, allowing for the simultaneous investigation of thermal and structural behavior. The governing equations for this analysis include the heat transfer equation and structural equations shown in Equation 1 and 2.

$$\rho c_p . \frac{\partial T}{\partial t} = \nabla . (k \nabla T) + Q \tag{1}$$

$$\nabla . \sigma + f_b = \rho . \frac{\partial^2 u}{\partial t^2} \tag{2}$$

Where $\rho$ is density, $c_p$ is the specific heat, $T$ is the temperature, $k$ is the thermal conductivity, $Q$ is the internal heat generation per unit volume, $\sigma$ is the stress tensor, $f_b$ is the body force per unit volume, and $u$ is the displacement vector.

The simulation approach involves the gradual addition of material layers controlled by element activation, with each step reflecting a phase in the deposition process where specific components are activated. This sequential deposition simulation monitors thermal and mechanical changes in each phase to provide a realistic depiction of the process. Simulations of convection and radiation are used to estimate heat transfer between the deposited material, the tool, and the environment, which influences cooling rates, temperature gradients, and potential distortions. Loading conditions, which include heat generated by friction, pressure, and shear forces (both longitudinal and rotational), have a significant impact on material flow, bonding, and final geometry. The finite element formulation breaks down the domain into smaller elements, which are activated and deactivated to precisely reflect the deposition process. Finite element equations are constructed from governing differential equations, with interpolation functions used to estimate the field variables (temperature, displacement) within each element. An implicit temporal integration approach is widely employed to provide stability in coupled analyses, and adequate boundary and beginning conditions are required for accurate simulation.

The equations involved in the activation and deactivation process include the activation function $A_c(t)$, which determines if an element is active or inactive at a given time as shown in Equation 3.

$$A_c(t) = \begin{cases} 1 & \text{if element } e \text{ is active at time } t. \\ 0 & \text{if element } e \text{ is inactive at time } t. \end{cases} \tag{3}$$

The heat transfer equation for active elements is modified to include the activation function and heat input due to deposition is depicted in Equation 4.

$$\rho c_p . \frac{\partial T}{\partial t} = \nabla.(k\nabla T) + Q + A_c(t).H(t) \qquad (4)$$

Where $H(t)$ is the heat input modulated by $A_c(t)$.

The structural response in active elements is represented by the Equation 5.

$$\nabla.(\sigma.A_c(t)) + f_b = \rho.\frac{\partial^2 u}{\partial t^2} \qquad (5)$$

The numerical model's use of the element activation and deactivation technique allows the simulation to accurately represent the dynamic character of the AFSD process. This method allows for predicting material properties, possible flaws, and the overall performance of the deposited structures by offering insightful information about the thermal and mechanical behavior of the deposited material as shown in Figure 6-10.

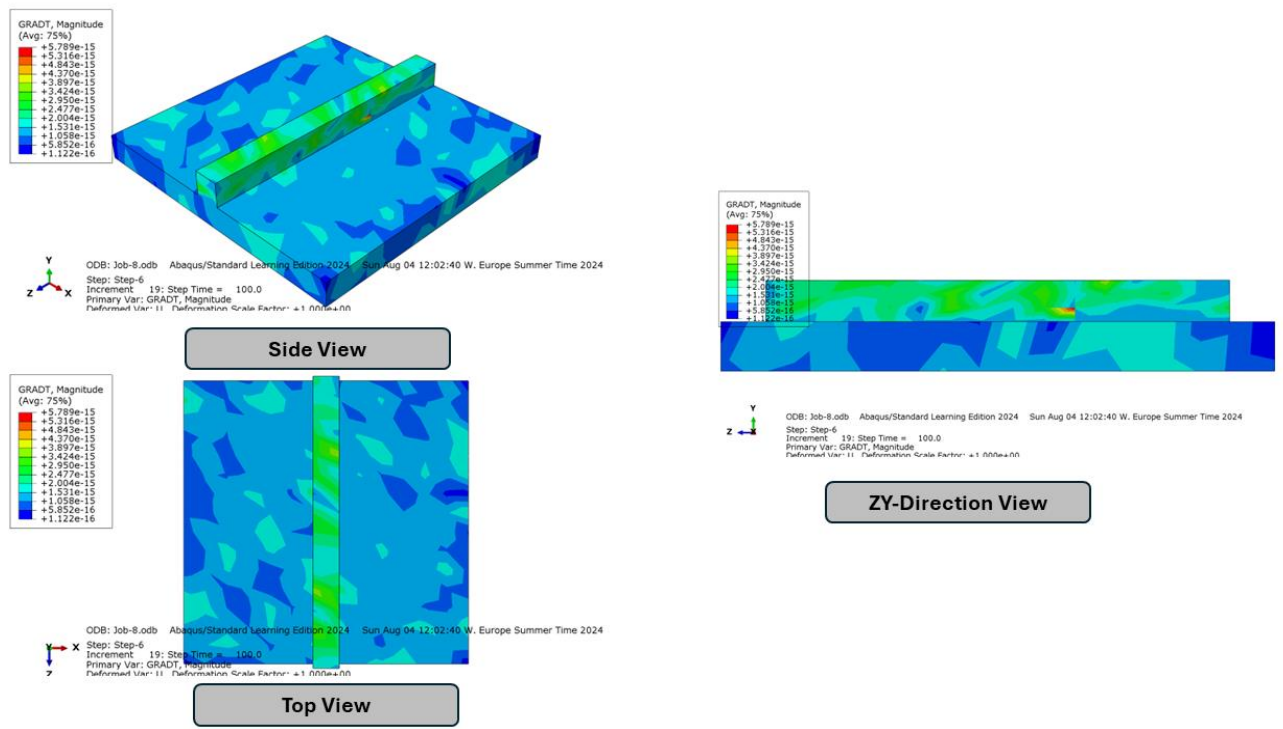

a)

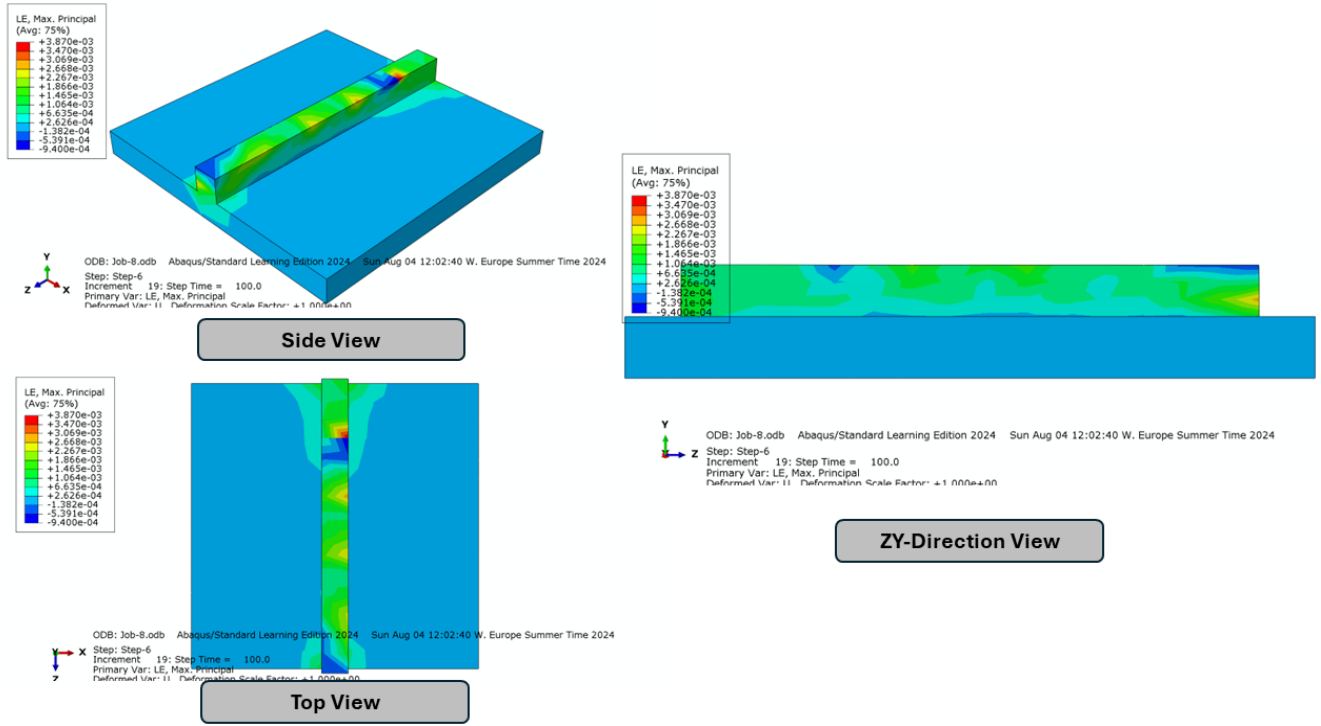

b)

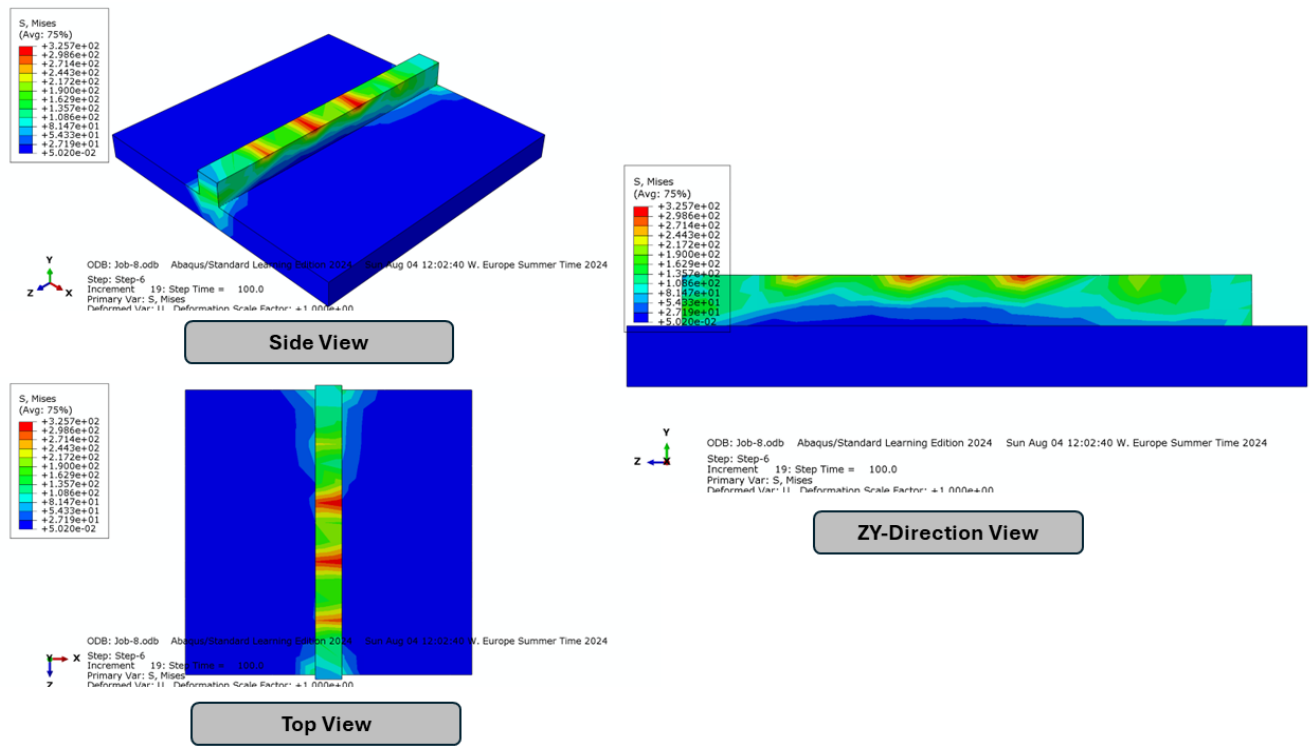

c)

**Figure 6.** Visualizations for Additive Friction Stir Deposited AA2024 walled structures a) GRADT, b) LE, and c) Von misses stress

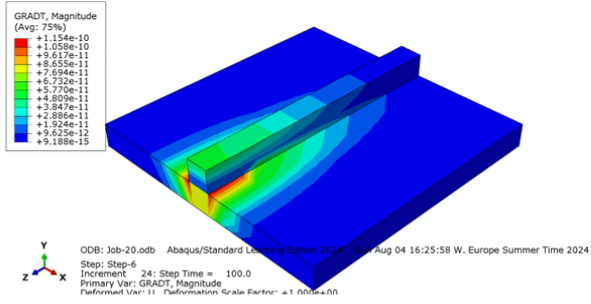

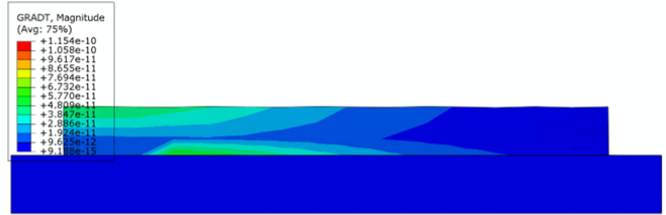

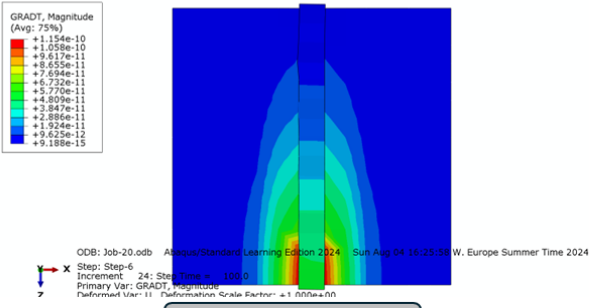

a)

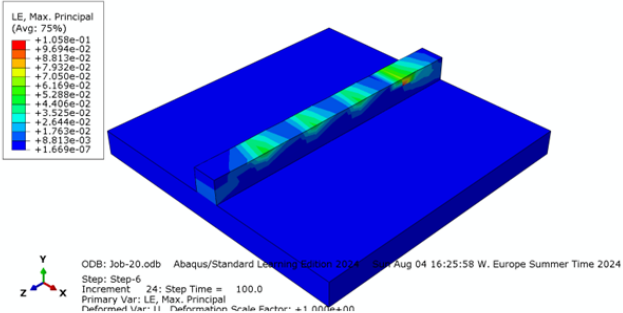

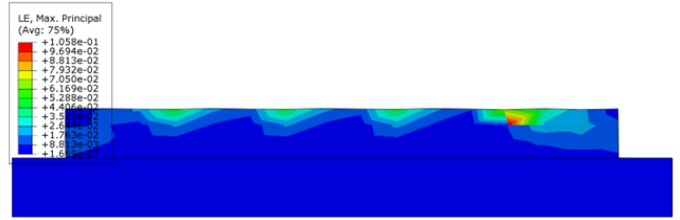

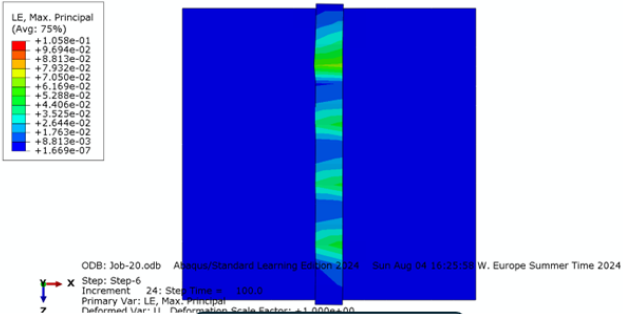

b)

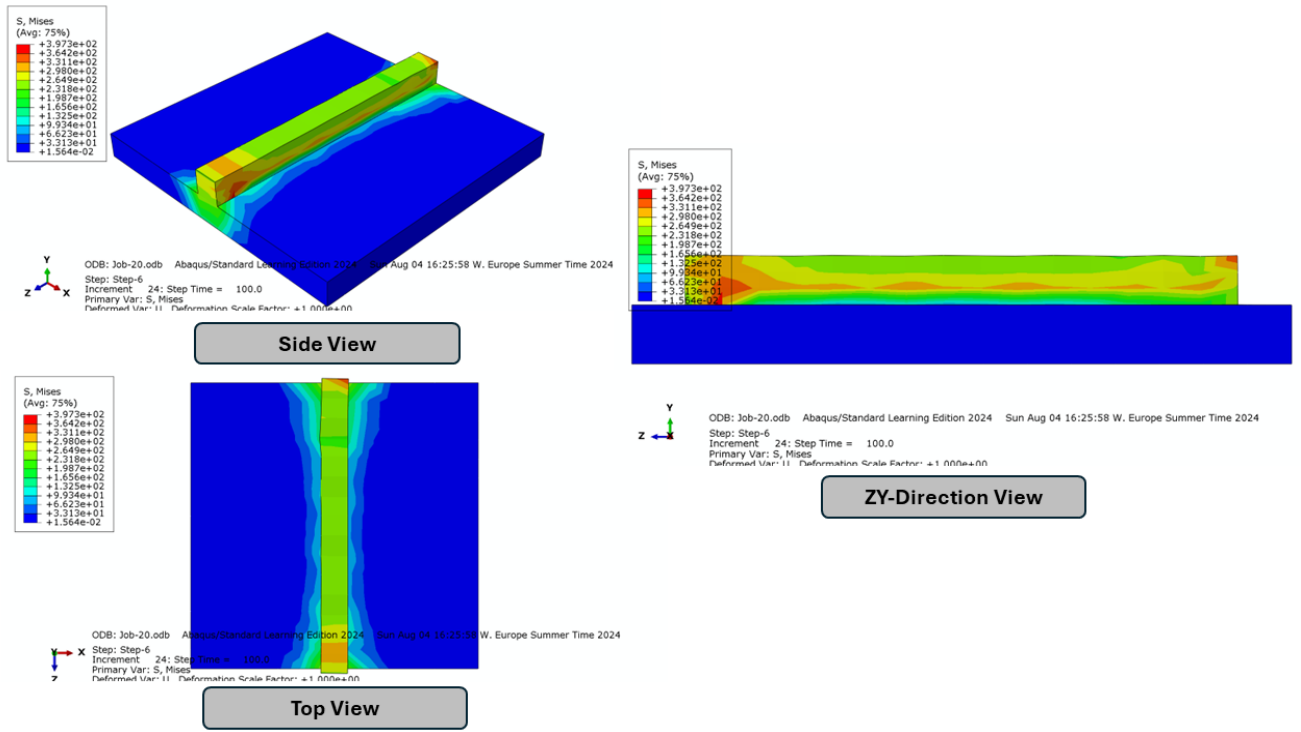

c)

**Figure 7.** Visualizations for Additive Friction Stir Deposited AA5086 walled structures a) GRADT, b) LE, and c) Von misses stress

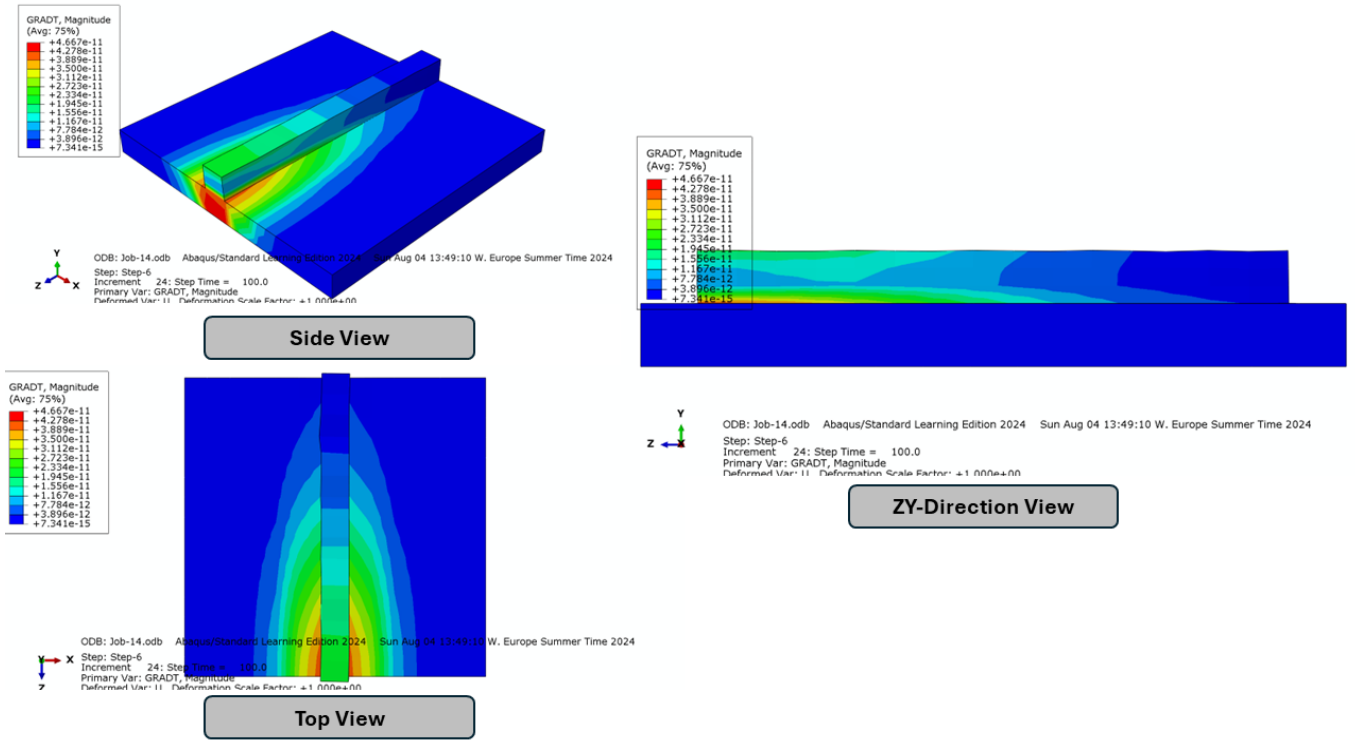

a)

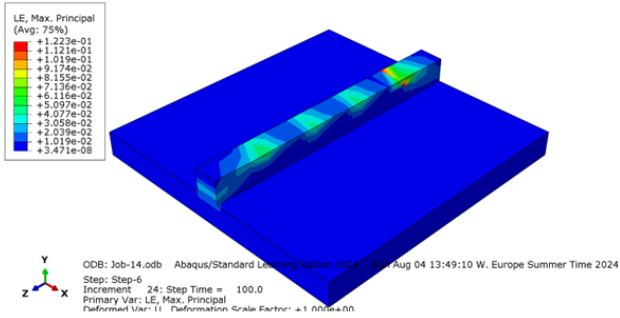

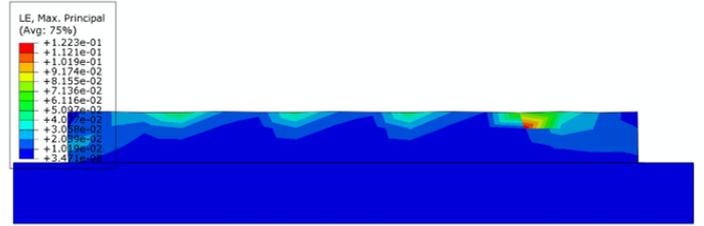

**Side View**

**ZY-Direction View**

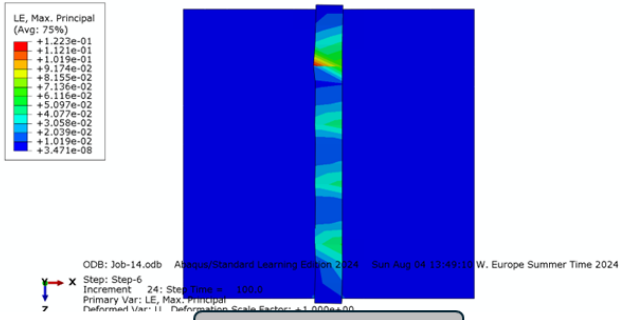

**Top View**

b)

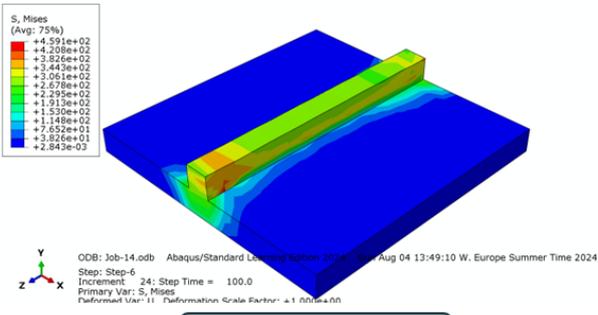

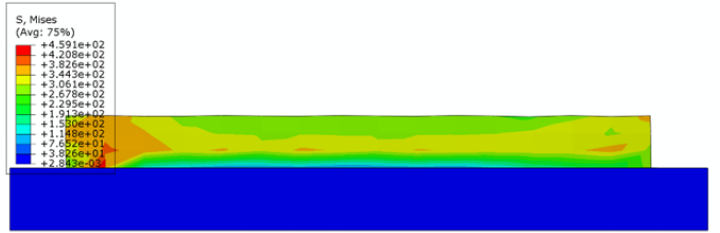

**Side View**

**ZY-Direction View**

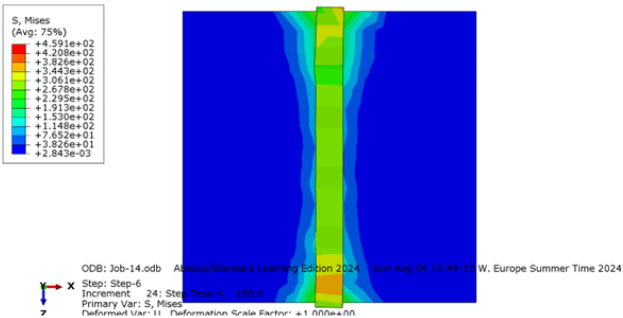

**Top View**

c)

**Figure 8.** Visualizations for Additive Friction Stir Deposited AA6061 walled structures a) GRADT, b) LE, and c) Von misses stress

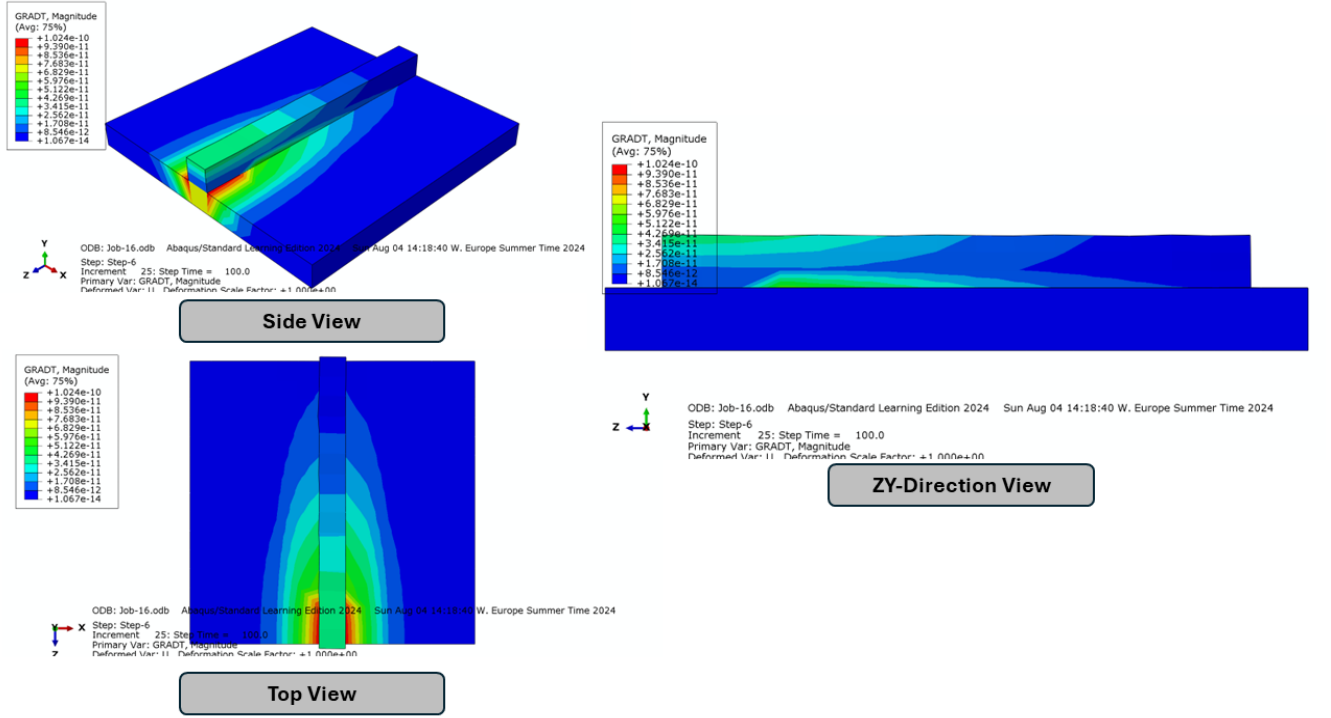

a)

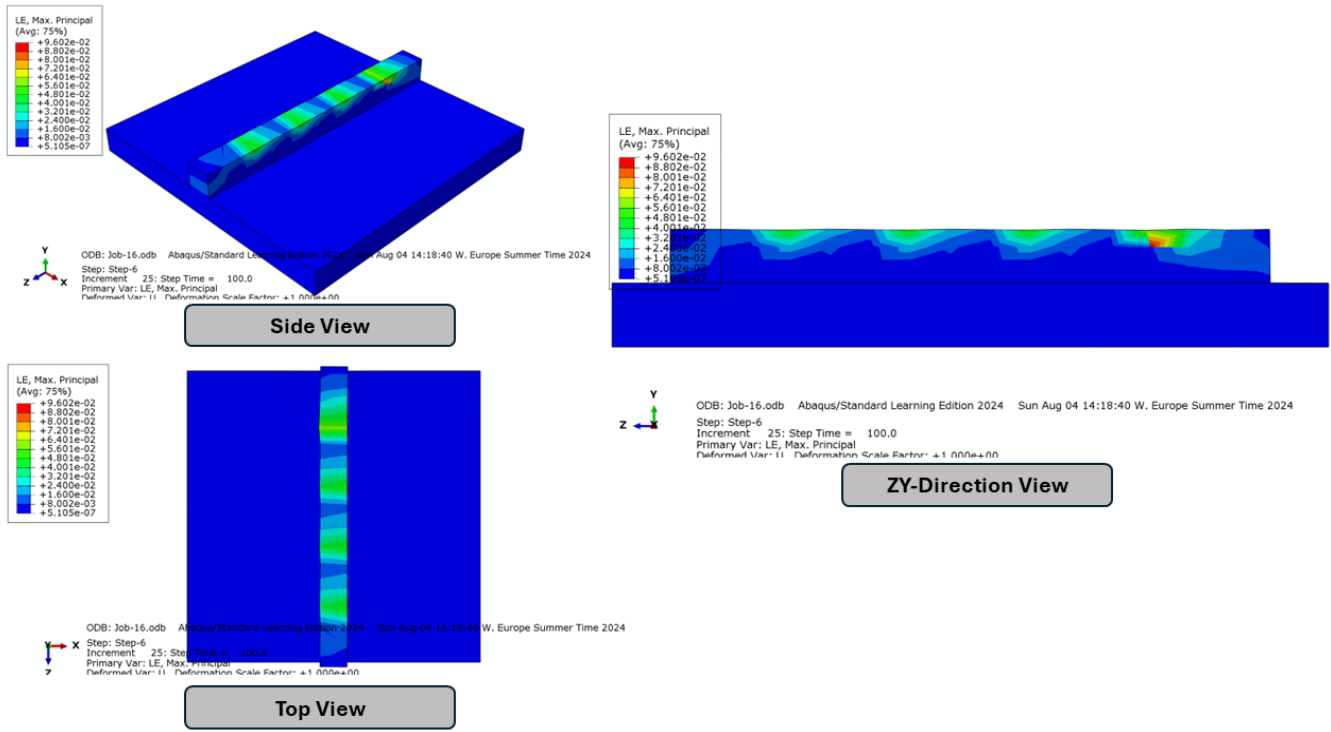

b)

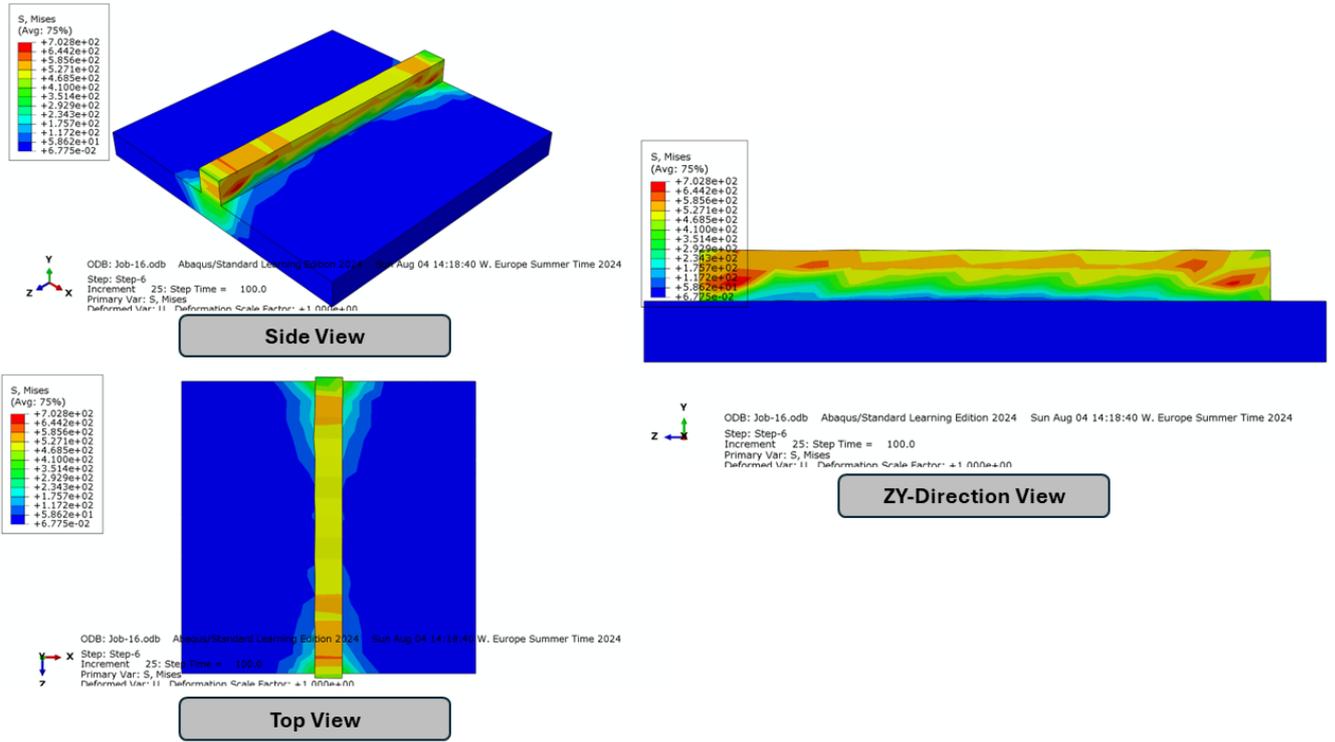

c)

**Figure 9.** Visualizations for Additive Friction Stir Deposited AA7075 walled structures a) GRADT, b) LE, and c) Von misses stress

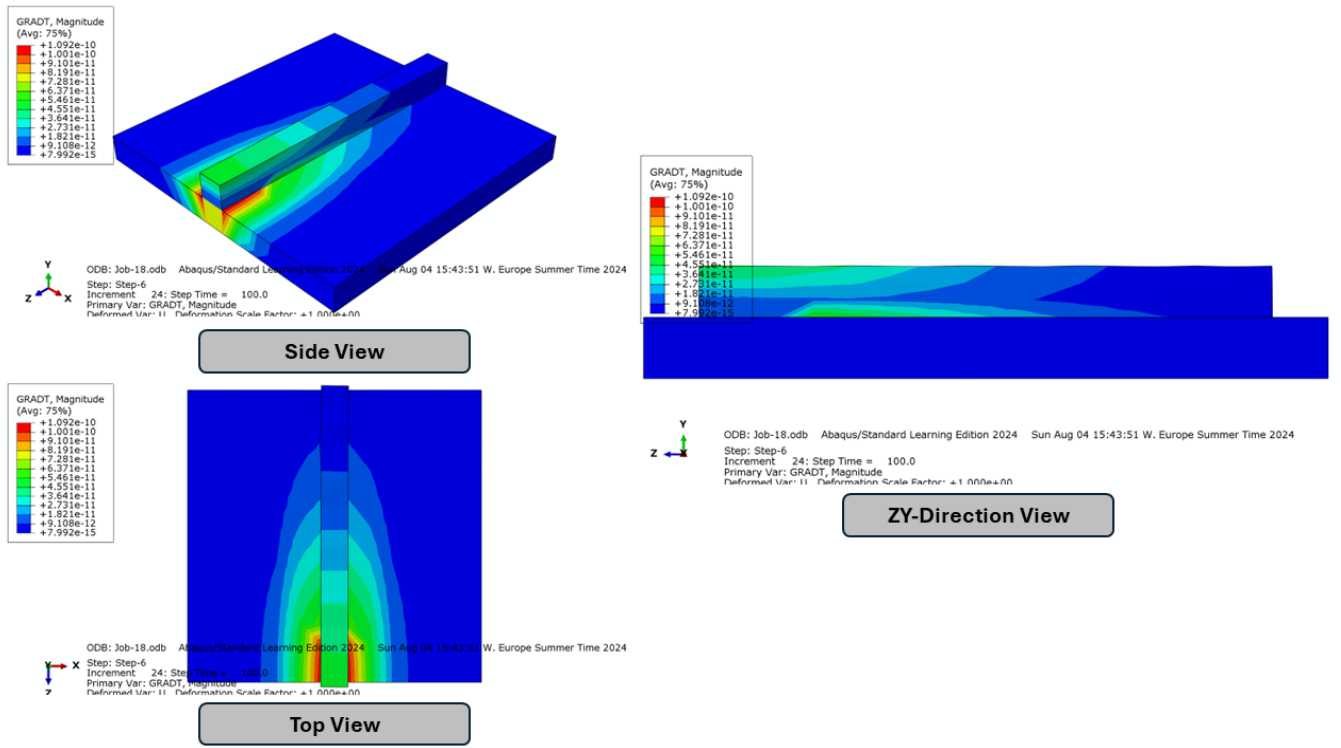

a)

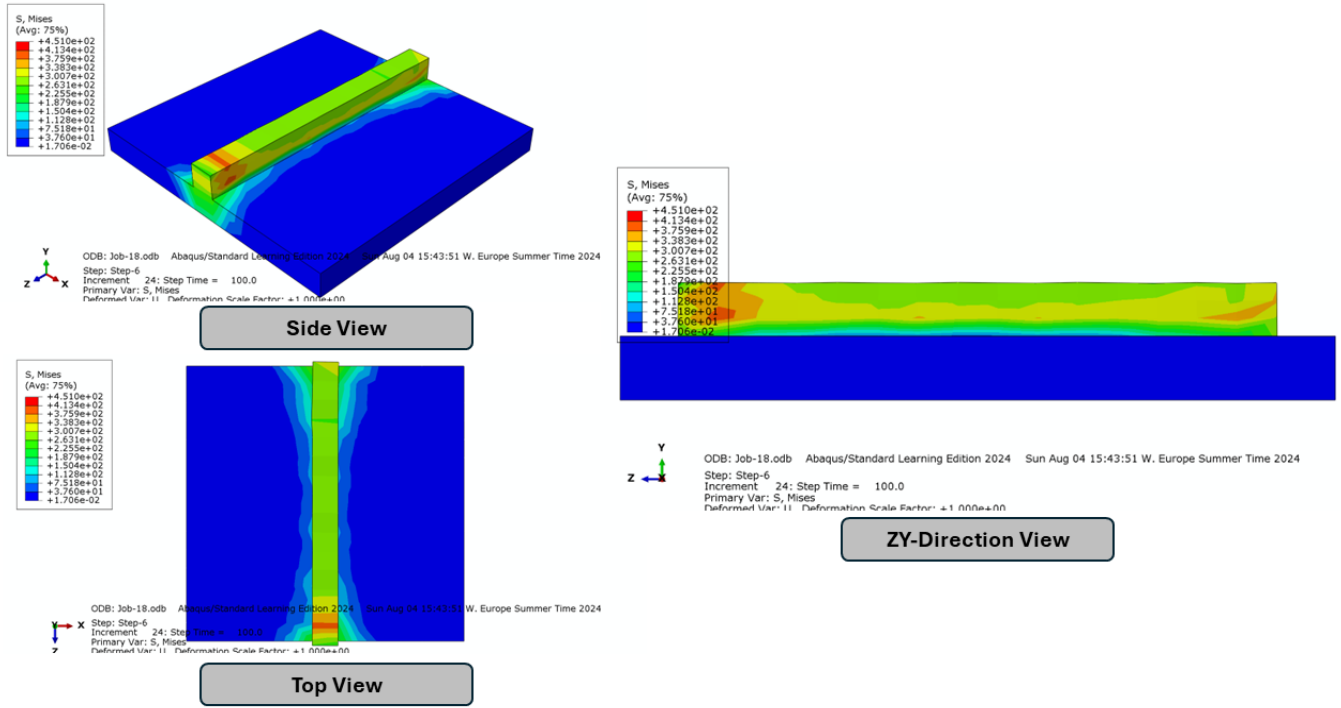

b)

**Figure 10.** Visualizations for Additive Friction Stir Deposited AA7075 walled structures a) GRADT, b) Von misses stress

### 4.2 Prediction of Von misses stress and Logarithmic Strain of additive friction stir deposited walled structures using Genetic algorithm coupled ML algorithms

Von Mises stress and Logarithmic Strain (LE) play critical roles in the Additive Friction Stir Deposition (AFSD) process. Von Mises stress is an important criterion for estimating the start of yielding in materials under complex loading situations. This is especially important in AFSD, where severe plastic deformation, high strain rates, and increased temperatures are involved. Von Mises stress determines whether the material will yield under applied pressures during deposition, which is critical for maintaining good bonding between layers without material failure. It also offers information about the ductility and material flow required for the specified bonding quality. In order to guarantee uniform deposition and reduce defects, we can improve process parameters such tool rotation speed, feed rate, and applied force by studying Von Mises stress.

Logarithmic Strain (LE), also known as True Strain, is equally significant in the AFSD process due to the large deformations involved. LE offers a precise measure of the deformation experienced by the material, which is crucial in processes involving extensive plastic deformation. Understanding the strain distribution helps identify regions of high deformation, critical for ensuring the structural integrity of the deposited material. LE also aids in accurately characterizing the material behavior under the high strains and temperatures typical of AFSD. This understanding is essential for optimizing the process and achieving the desired material properties.

In this study, we developed and compared reliable and efficient coupled algorithms that combine Decision Tree Regression and Random Forest Regression with Genetic Algorithm (GA) optimization to forecast Von Mises stress and Logarithmic Strain (LE) in the Additive Friction Stir Deposition (AFSD) process. The input parameters include the elastic modulus, specific heat, shear translation, shear rotation, and heat source properties of the alloys utilized. These methods combine the predictive accuracy of Decision Tree and Random Forest regressions with GA's optimization capabilities, resulting in optimal model performance for complicated material behavior prediction in AFSD.

Decision Tree Regression is a non-parametric supervised learning technique for regression tasks. It creates a tree with nodes representing features (or attributes), branches representing decision rules, and leaves representing outcomes. The goal is to build a model that can predict the value of a target variable using basic decision rules derived from data attributes. The hyperparameters of a Decision Tree model are the maximum depth of the tree ($d$), minimum number of samples required to split an internal node ($s$), and minimum number of samples required to be at a leaf node ($l$). Random Forest Regression extends the decision tree model by building numerous decision trees during training and calculating the average prediction of each tree. This method improves prediction performance by decreasing overfitting and increasing generalization. The important hyperparameters for a Random Forest model are the number of estimators ($n$), maximum depth ($d$), minimum sample split ($s$), and minimum sample leaf ($l$). The prediction of a Random Forest model for an input $X$ is the average prediction of all the trees in the forest as shown in Equation 6.

$$\hat{y} = \frac{1}{n}\sum_{i=1}^{n} h_i(X) \tag{6}$$

Where $h_i(X)$ is the prediction of the *i-th* tree.

Each individual in the GA population represents a set of hyperparameters shown in Table 2. For a Decision Tree, an individual might be represented as ($d, s, l$) while for a Random Forest, it might be ($n, d, s, l$). The fitness function evaluates the performance of the model with the given hyperparameters. Equation 7 and 8 shows the calculation of fitness function for decision tree and random forest algorithm.

$$Fitness(individual) = \frac{1}{MSE(y_{test}, h(X_{test}))} \tag{7}$$

$$Fitness\ (individual) = \frac{1}{MSE(y_{test}, \hat{y}\ (X_{test}))} \tag{8}$$

In each generation, the GA selects individuals based on their fitness scores, creates new offspring through crossover, and introduces mutations to maintain diversity. This iterative process continues until a predetermined number of generations is completed or convergence is achieved as shown in Figure 11 and 12.

**Table 2.** Genetic Algorithm Parameters

| Population size | Generations | Crossover probability | Mutation probability |
|---|---|---|---|
| 50 | 200 | 0.8 | 0.1 |

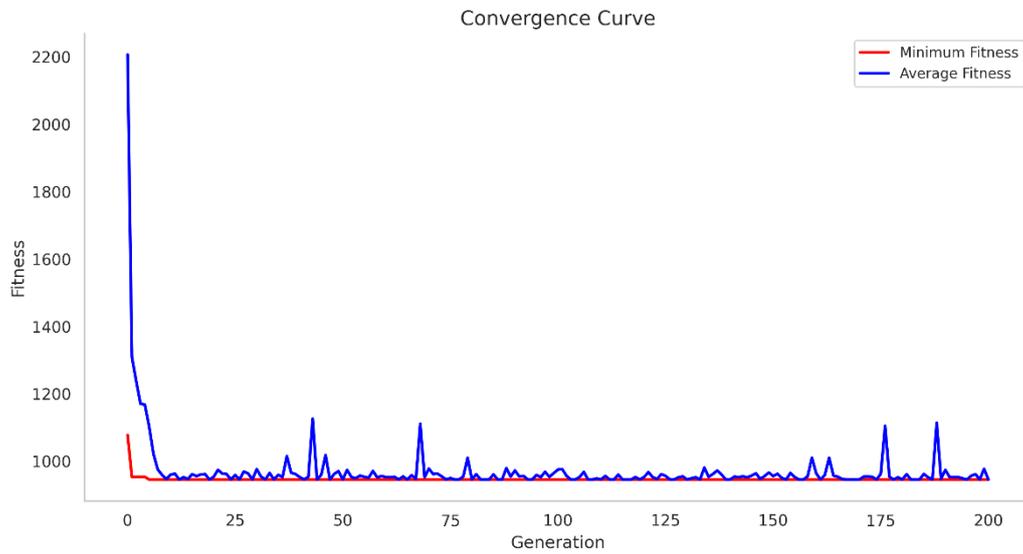

s)

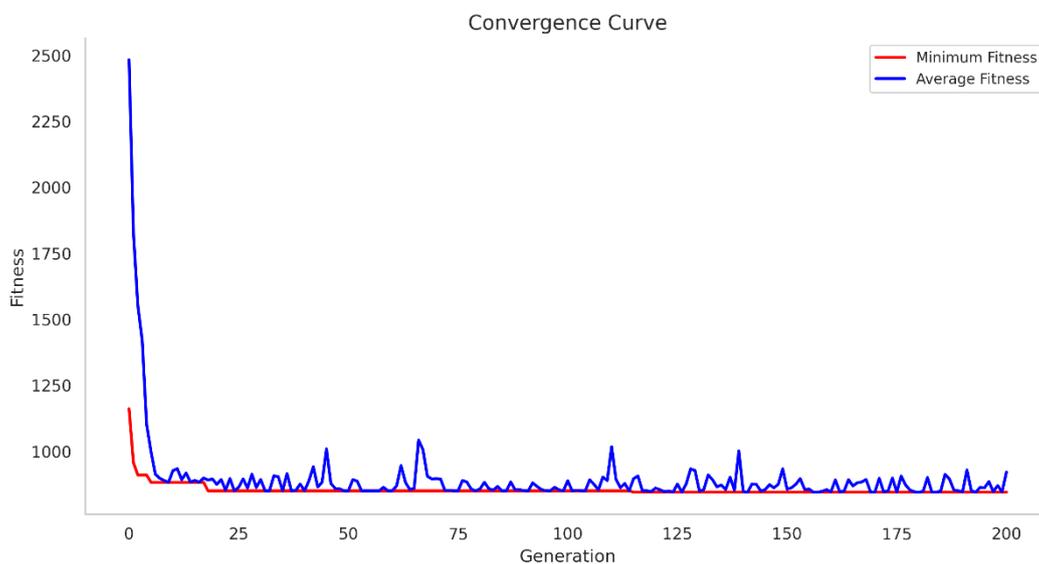

b)

**Figure 11.** Convergence curves of a) GA-DT, b) GA-RF for predicting the von misses strain of additive friction stir deposited aluminium based walled structures

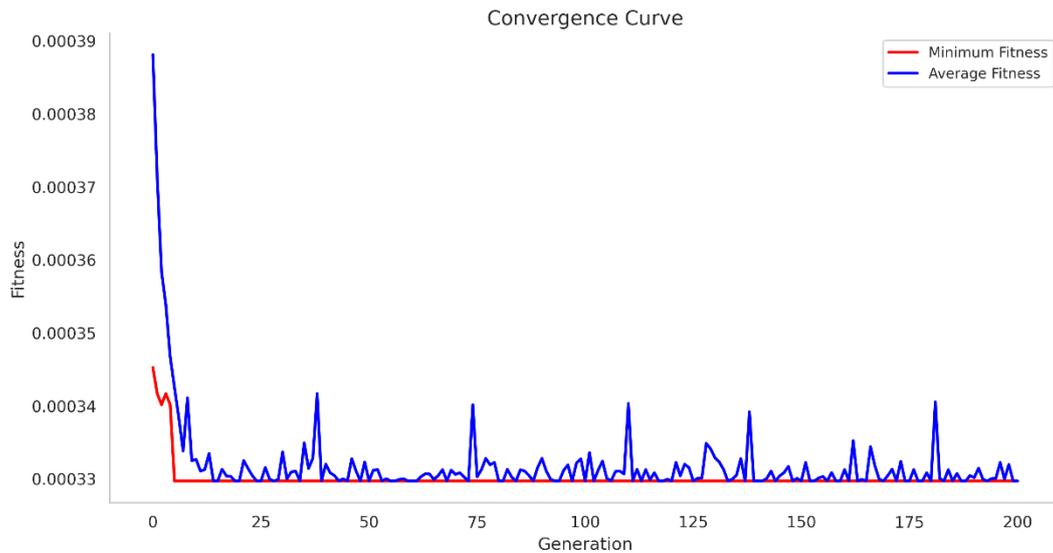

a)

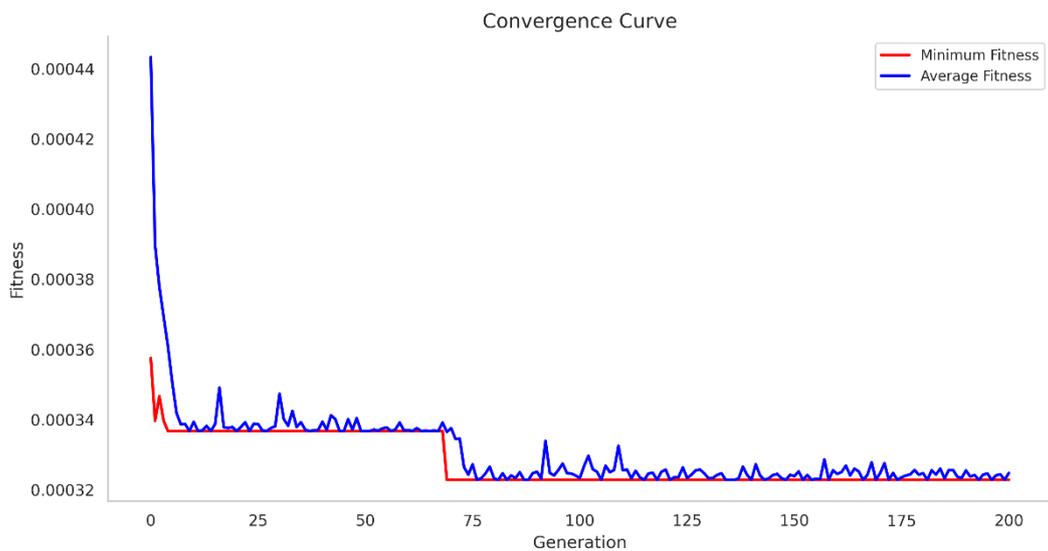

b)

**Figure 12**. Convergence curves of a) GA-DT, b) GA-RF for predicting the logarithmic strain of additive friction stir deposited aluminium based walled structures

In Figure 11, we can observe the optimization process for predicting von Mises stress. Both GA-DT (11a) and GA-RF (11b) models show a rapid initial improvement in fitness scores, followed by a more gradual optimization as the number of generations increases. The GA-RF model appears to achieve slightly higher fitness scores compared to GA-DT, suggesting it may be more effective in predicting von Mises stress for this application.

Figure 12 presents similar convergence curves for the prediction of logarithmic strain. Again, both GA-DT (12a) and GA-RF (12b) models demonstrate quick initial improvements in fitness scores. However, the convergence pattern for logarithmic strain prediction seems to be more erratic compared to von Mises stress prediction, particularly for the GA-RF model. This could indicate that predicting logarithmic strain is a more challenging task, possibly due to its higher sensitivity to local variations in the material properties or process parameters.

Tables 3 and 4 present the best hyperparameters obtained through genetic algorithm optimization for the Decision Tree (GA-DT) and Random Forest (GA-RF) models when predicting von Mises stress and logarithmic strain, respectively, in additive friction stir deposited aluminum-based walled structures. From Table 3 its observed that the GA-DT model favors a deeper tree with a maximum depth of 10, while the GA-RF model uses shallower trees with a maximum depth of 5. This suggests that for von Mises stress, the Decision Tree benefits from more complex decision paths, while the Random Forest achieves better results with simpler individual trees but implement the power of ensemble learning. Both models prefer a small minimum number of samples to split an internal node (2 for GA-DT, 3 for GA-RF) and the minimum possible number of samples at a leaf node (1 for both). This indicates that the models benefit from fine-grained decision-making capabilities. The GA-RF model uses 93 estimators (trees), which is a relatively high number, suggesting that the ensemble's diversity contributes significantly to its predictive power for von Mises stress. From Table 4 it is observed that the trend is reversed for tree depth, with GA-DT using shallower trees (depth 5) and GA-RF using deeper trees (depth 11). This implies that logarithmic strain prediction benefits from different model architectures compared to von Mises stress prediction. Both models maintain the preference for a small minimum number of samples to split (2) and minimum samples at leaf nodes (1), similar to the von Mises stress prediction. The GA-RF model for logarithmic strain uses fewer estimators (23) compared to the von Mises stress model, suggesting that fewer, more complex trees are more effective for this particular prediction task.

**Table 3.** Obtained best parameters for predicting the von misses stress of additive friction stir deposited aluminium based walled structures

| Algorithms | Best Max Depth | Best Min Samples Split | Best Min Samples Leaf | Best N Estimators |
|---|---|---|---|---|
| GA-DT | 10 | 2 | 1 | - |
| GA-RF | 5 | 3 | 1 | 93 |

**Table 4.** Obtained best parameters for predicting the logarithmic strain of additive friction stir deposited aluminium based walled structures

| Algorithms | Best Max Depth | Best Min Samples Split | Best Min Samples Leaf | Best N Estimators |
|---|---|---|---|---|
| GA-DT | 5 | 2 | 1 | - |
| GA-RF | 11 | 2 | 1 | 23 |

Tables 5 and 6, along with Figures 13 and 14, provide a comprehensive view of the performance metrics for the GA-DT and GA-RF models in predicting von Mises stress and logarithmic strain, respectively, for additive friction stir deposited aluminum-based walled structures.

**Table 5.** Metric features obtained for predicting the von misses stress of additive friction stir deposited aluminium based walled structures

| Algorithms | RMSE | MAE | R² Value |
|---|---|---|---|
| GA-DT | 30.75 | 22.75 | 0.9638 |
| GA-RF | 29.08 | 23.20 | 0.9676 |

**Table 6.** Metric features obtained for predicting the logarithmic strain of additive friction stir deposited aluminium based walled structures

| Algorithms | RMSE | MAE | R² Value |
|---|---|---|---|
| GA-DT | 0.017 | 0.010 | 0.7142 |
| GA-RF | 0.017 | 0.011 | 0.7201 |

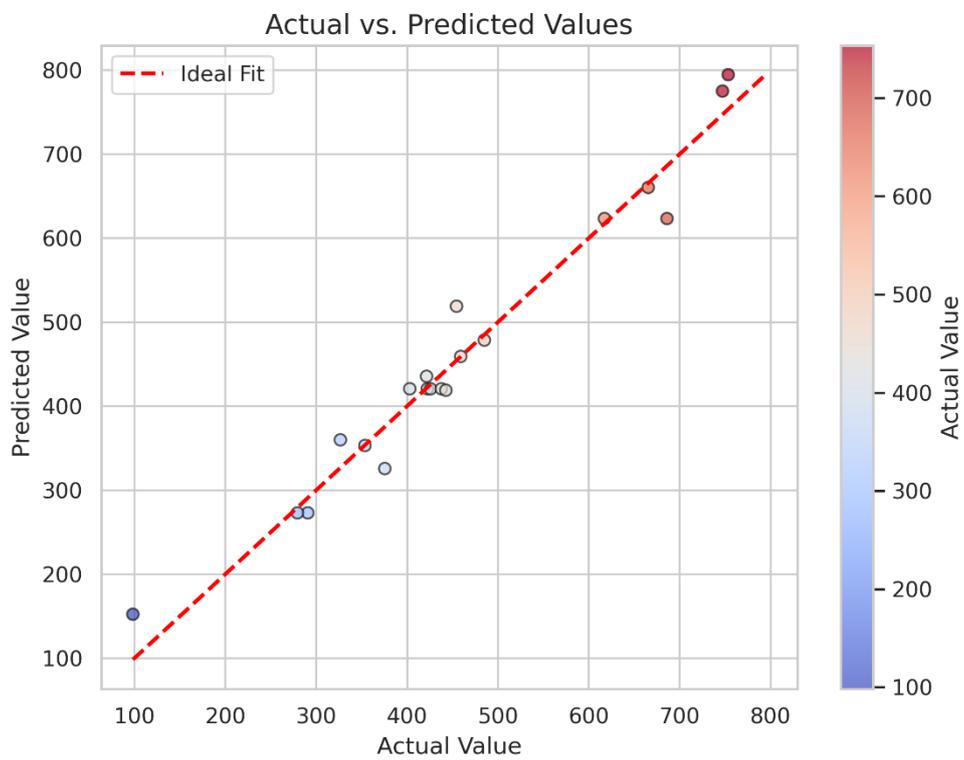

a)

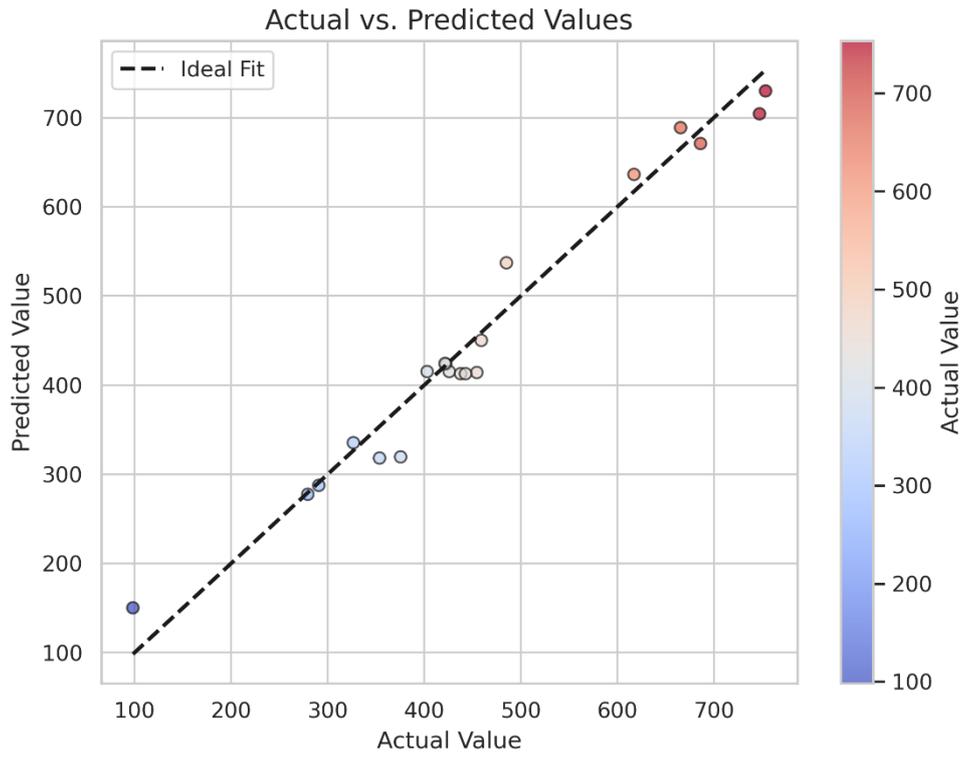

b)

**Figure 13.** Actual vs predicted values plots of a) GA-DT, b) GA-RF for predicting the von misses stress of additive friction stir deposited aluminium based walled structures

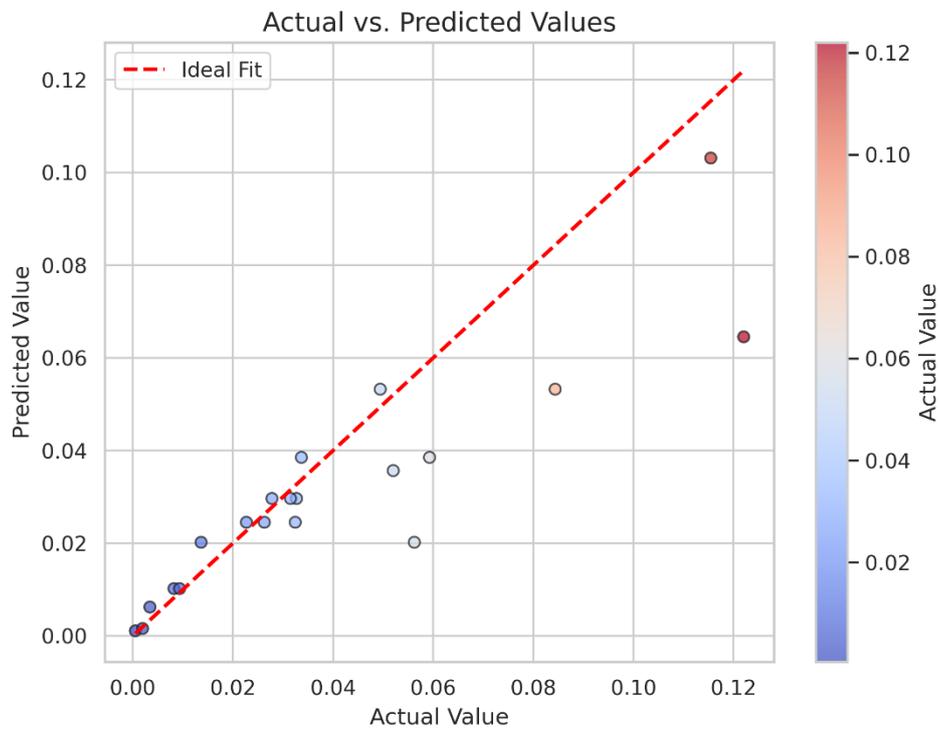

a)

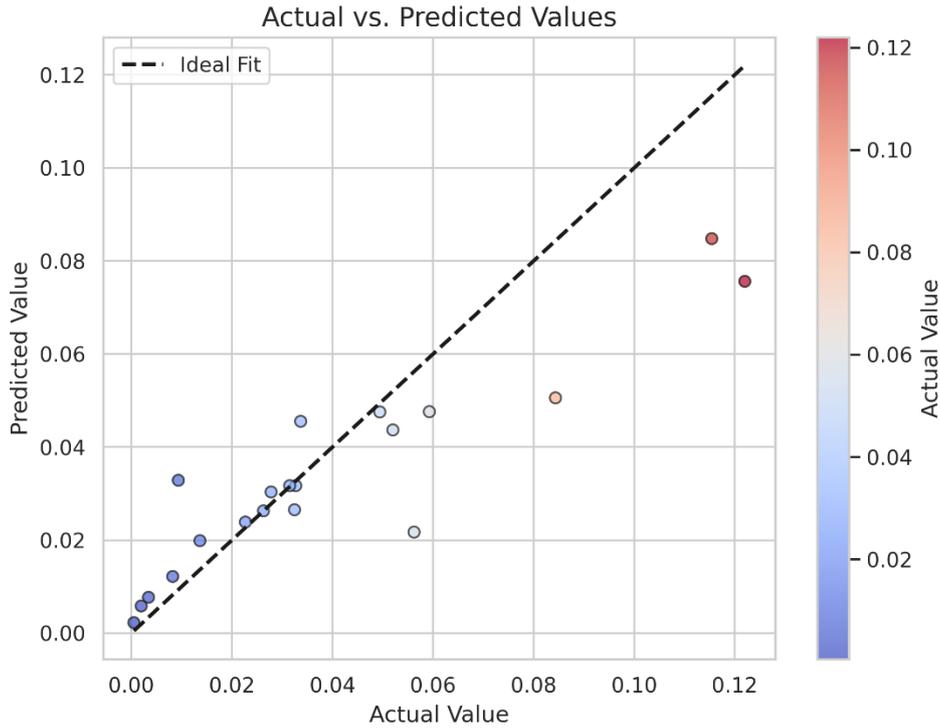

b)

**Figure 14.** Actual vs predicted values plots of a) GA-DT, b) GA-RF for predicting the logarithmic strain of additive friction stir deposited aluminium based walled structures

For von Mises stress prediction (Table 5 and Figure 13), both models demonstrate excellent performance. The GA-RF model slightly outperforms GA-DT, with lower RMSE (29.08 vs 30.75), similar MAE, and a higher R² value (0.9676 vs 0.9638). This superior performance is visually confirmed in Figure 13, where both models show a strong correlation between predicted and actual values, with GA-RF (13b) displaying a slightly tighter clustering along the ideal prediction line compared to GA-DT (13a). In contrast, the logarithmic strain prediction (Table 6 and Figure 14) shows lower overall performance for both models, but with GA-RF still marginally outperforming GA-DT. Both models have identical RMSE (0.017), with GA-DT showing a slightly lower MAE (0.010 vs 0.011) and GA-RF achieving a marginally higher R² value (0.7201 vs 0.7142). Figure 14 visually corroborates these results, showing more scattered predictions for both models compared to the von Mises stress predictions, with GA-RF (14b) displaying a slightly better fit than GA-DT (14a).

The significant difference in prediction performance between von Mises stress and logarithmic strain indicates that the latter is a more difficult quantity to forecast in the AFSD process. This could be because logarithmic strain is more sensitive to local fluctuations in material qualities or process parameters, or it could be influenced by factors that are not fully captured by the current set of input variables. Despite this, both GA-DT and GA-RF models indicate the ability to capture underlying patterns in data, with GA-RF consistently outperforming GA-DT in both prediction tasks. These findings show the promise of machine learning techniques, particularly those optimized with genetic algorithms, for predicting complicated material behaviors in advanced manufacturing processes such as AFSD.

## 5. Conclusion

In conclusion, this study has successfully developed a novel biomimetic machine learning approach for predicting the mechanical properties of Additive Friction Stir Deposited (AFSD) aluminum alloy walled structures. By integrating finite element analysis with genetic algorithm-optimized Decision Tree and Random Forest models, the study demonstrated the ability to accurately predict von Mises stress and logarithmic strain across five aluminum alloys: AA2024, AA5083, AA5086, AA7075, and AA6061. The GA-RF model, in particular, showed superior performance with $R^2$ values of 0.9676 for von Mises stress and 0.7201 for logarithmic strain prediction. These results highlight the potential of developed approach to significantly enhance process optimization and quality control in AFSD manufacturing. This study opens up a number of promising options for further research. First, the technology might be expanded to encompass a broader spectrum of aluminum alloys and other materials appropriate for AFSD. Second, combining real-time process monitoring data into machine learning models may increase prediction accuracy and allow for adaptive control measures. Third, integrating this approach with other modern manufacturing techniques may result in hybrid processes with expanded capabilities. Finally, creating a user-friendly interface for this predictive tool could help it gain traction in industrial settings, perhaps transforming the field of additive manufacturing for high-performance aluminum structures.

## References

1. Yu, H.Z. and Mishra, R.S., 2021. Additive friction stir deposition: a deformation processing route to metal additive manufacturing. Materials Research Letters, 9(2), pp.71-83.
2. Patil, S.M., Sharma, S., Joshi, S.S., Jin, Y., Radhakrishnan, M. and Dahotre, N.B., 2024. Additive friction stir deposition of Al 6061-B4C composites: Process parameters, microstructure and property correlation. Materials Science and Engineering: A, 910, p.146840.
3. KORGANCI, M. and BOZKURT, Y., 2024. Recent developments in Additive Friction Stir Deposition (AFSD). Journal of Materials Research and Technology.
4. Liu, H., Xu, M. and Li, X., 2024. Achievement of high-reliability and high-efficient deposit of PA66 by additive friction stir deposition. Composites Part B: Engineering, 284, p.111682.
5. Chen, L., Lu, L., Zhu, L., Yang, Z., Zhou, W., Ren, X. and Zhang, X., 2024. Microstructure evolution and mechanical properties of multilayer AA6061 alloy fabricated by additive friction stir deposition. Metallurgical and Materials Transactions A, 55(4), pp.1049-1064.

6. Liu, H., Liu, Y., Liang, T., Xie, R., Liu, B., Wang, Z., Han, Y. and Chen, S., 2024. Effect of press depth on defect formation in friction-rolling additive manufacturing. Journal of Manufacturing Processes, 119, pp.305-320.

7. Li, X., Li, X., Hu, S., Liu, Y. and Ma, D., 2024. Additive friction stir deposition: a review on processes, parameters, characteristics, and applications. The International Journal of Advanced Manufacturing Technology, pp.1-18.

8. Xu, X., Qiu, W., Wan, D., Wu, J., Zhao, F. and Xiong, Y., 2024. Numerical modelling of the viscoelastic polymer melt flow in material extrusion additive manufacturing. Virtual and Physical Prototyping, 19(1), p.e2300666.

9. Liu, F.C., Feng, A.H., Pei, X., Hovanski, Y., Mishra, R.S. and Ma, Z.Y., 2024. Friction stir based welding, processing, extrusion and additive manufacturing. Progress in Materials Science, p.101330.

10. Patel, A. and Taufik, M., 2024. Extrusion-based technology in additive manufacturing: a comprehensive review. Arabian Journal for Science and Engineering, 49(2), pp.1309-1342.

11. Krishnanand, Singh, V., Mittal, V., Branwal, A.K., Sharma, K. and Taufik, M., 2024. Extrusion strategies in fused deposition additive manufacturing: A review. Proceedings of the Institution of Mechanical Engineers, Part E: Journal of Process Mechanical Engineering, 238(2), pp.988-1012.

12. Maurya, M., Maurya, A. and Kumar, S., 2024. Variants of friction stir based processes: review on process fundamentals, material attributes and mechanical properties. Materials Testing, 66(2), pp.271-287.

13. Gottwald, R.B., Gotawala, N., Erb, D.J. and Hang, Z.Y., 2024. An exploratory study on miniaturized additive friction stir deposition. Journal of Manufacturing Processes, 126, pp.154-164.

14. Rao, A.G. and Meena, N., 2024. Friction Stir-Based Additive Manufacturing. Friction Stir Welding and Processing: Fundamentals to Advancements, pp.293-305.

15. Qiao, Q., Liu, Q., Pu, J., Shi, H., Li, W., Zhu, Z., Guo, D., Qian, H., Zhang, D., Li, X. and Kwok, C.T., 2024. A comparative study of machine learning in predicting the mechanical properties of the deposited AA6061 alloys via additive friction stir deposition. Materials Genome Engineering Advances, 2(1), p.e31.

16. Zhu, Y., Wu, X., Gotawala, N., Higdon, D.M. and Hang, Z.Y., 2024. Thermal prediction of additive friction stir deposition through Bayesian learning-enabled explainable artificial intelligence. Journal of Manufacturing Systems, 72, pp.1-15.

17. Shi, T., Wu, J., Ma, M., Charles, E. and Schmitz, T., 2024. AFSD-Nets: A Physics-informed machine learning model for predicting the temperature evolution during additive friction stir deposition. Journal of Manufacturing Science and Engineering, 146(8).